\documentclass{article}

\PassOptionsToPackage{numbers, compress}{natbib}
\usepackage[preprint]{neurips_2025} 

\newcommand{\eat}[1]{}
\usepackage[utf8]{inputenc} 
\usepackage[T1]{fontenc}    
\usepackage{hyperref}       
\usepackage{url}            
\usepackage{booktabs}       
\usepackage{amsfonts}       
\usepackage{nicefrac}       
\usepackage{microtype}      
\usepackage{xcolor}         

\usepackage{pifont}
\newcommand{\cmark}{\ding{52}}  %
\newcommand{\xmark}{\ding{56}}  %
\usepackage{graphicx}
\usepackage{enumitem}
\usepackage[color,matrix,arrow,all]{xy}

\usepackage{url}
\usepackage{listings}
\usepackage[flushleft]{threeparttable}
\usepackage{makecell}
\usepackage{xparse}
\usepackage{wrapfig}
\usepackage{tcolorbox}

\usepackage{epsfig}
\usepackage{multirow}

\usepackage{bbm}

\definecolor{shadecolor}{RGB}{220,220,220}

\definecolor{inputcolor}{RGB}{255,139,35}
\definecolor{outputcolor}{RGB}{120,212,252}
\definecolor{embedcolor}{RGB}{254,127,156}
\definecolor{maskcolor}{RGB}{122,128,255}
\definecolor{ecolor}{RGB}{58,149,54}

\definecolor{highcolor}{RGB}{255,153,153}
\definecolor{midcolor}{RGB}{255,204,204}
\definecolor{lowcolor}{RGB}{204,229,255}

\usepackage{tikz}
\usetikzlibrary{shapes,snakes}
\usetikzlibrary{calc}

\usepackage{algorithm} 
\usepackage{algorithmic}  
\usepackage[algo2e]{algorithm2e} 

\usepackage[export]{adjustbox}

\definecolor{green}{RGB}{0,128,0}

\definecolor{yellow}{RGB}{255,200,18}

\newcommand{\bi}{\begin{itemize}}
\newcommand{\ei}{\end{itemize}}

\newcommand{\be}{\begin{enumerate}}
\newcommand{\ee}{\end{enumerate}}
\newcommand{\beqn}{\begin{eqnarray*}}
\newcommand{\eeqn}{\end{eqnarray*}}

\newcommand{\ie}{\textit{i.e.,}\xspace}
\newcommand{\eg}{\textit{e.g.,}\xspace}

\usepackage{tcolorbox}
\usepackage{colortbl}
\usepackage{soul}
\definecolor{c1}{cmyk}{0,0.6175,0.8848,0.1490}
\definecolor{c2}{cmyk}{0.1127,0.6690,0,0.4431}
\definecolor{c3}{cmyk}{0.3081,0,0.7209,0.3255}
\definecolor{c4}{cmyk}{0.6765,0.2017,0,0.0667}
\definecolor{c5}{cmyk}{0,0.8765,0.7099,0.3647}

\newtcbox{\hlprimarytab}{on line, rounded corners, box align=base, colback=c3!10,colframe=white,size=fbox,arc=3pt, before upper=\strut, top=-2pt, bottom=-4pt, left=-2pt, right=-2pt, boxrule=0pt}
\newtcbox{\hlsecondarytab}{on line, box align=base, colback=red!10,colframe=white,size=fbox,arc=3pt, before upper=\strut, top=-2pt, bottom=-4pt, left=-2pt, right=-2pt, boxrule=0pt}
\newtcolorbox[]{finding}[0]{colback=gray!10, colframe=black, width=\columnwidth, boxrule=0.4pt, 
left=0mm, right=0mm, top=0mm, bottom=0mm, before skip=3pt, after skip=3pt, sharp corners}

\makeatletter
    \newcommand\figcaption{\def\@captype{figure}\caption}
    \newcommand\tabcaption{\def\@captype{table}\caption}
\makeatother

\tikzstyle{mybox} = [draw=black, fill=black!5, thick,
   rectangle, rounded corners, inner sep=0pt, inner ysep=6pt]
\tikzstyle{fancytitle} =[fill=black, text=white]

\NewDocumentCommand{\nan}{ mO{} }{\textcolor{blue}{\textsuperscript{\textit{Nan}}\textsf{\textbf{\small[#1]}}}}

\NewDocumentCommand{\yuyu}{ mO{} }{\textcolor{green}{\textsuperscript{\textit{Yuyu}}\textsf{\textbf{\small[#1]}}}}

\NewDocumentCommand{\yifan}{ mO{} }{\textcolor{brown}{\textsuperscript{\textit{Yifan}}\textsf{\textbf{\small[#1]}}}}

\newcommand{\ourRoPE}{{Unified RoPE}\xspace}

\newcommand{\model}{{TransXSSM}\xspace}

\usepackage{marginnote}
\let\oldmarginpar\marginpar
\renewcommand\marginpar[1]{\-\oldmarginpar[\raggedleft\footnotesize #1]%
	{\raggedright\footnotesize\color{blue} #1}} 
\usepackage{marginnote}
\let\oldmarginnote\marginnote
\renewcommand\marginnote[1]{\-\oldmarginnote[\raggedleft\footnotesize #1]%
	{\raggedright\footnotesize\color{blue} #1}} %

\usepackage[utf8]{inputenc} 
\usepackage[T1]{fontenc}    
\usepackage{hyperref}       
\usepackage{url}            
\usepackage{booktabs}       
\usepackage{amsfonts}       
\usepackage{nicefrac}       
\usepackage{microtype}      
\usepackage{xcolor}         
\usepackage{graphicx}       
\usepackage{amsmath}        
\usepackage{amsthm}         
\usepackage{tabularx}       
\usepackage{etoolbox}       
\usepackage{subcaption}     
\usepackage{wrapfig}        

\title{\model: A Hybrid Transformer–State Space Model with Unified Rotary Position Embedding}
\author{
  {Bingheng Wu}$^{1}$\quad
  {Jingze Shi}$^{1}$\quad
  {Yifan Wu}$^{1}$\quad \\
  \textbf{Nan Tang}$^{1}$\quad  
  \textbf{Yuyu Luo}$^{1}$ \\
  $^{1}$HKUST (Guangzhou) \quad
}

\begin{document}

\maketitle

\begin{abstract}

Transformers exhibit proficiency in capturing long-range dependencies, whereas State Space Models (SSMs) facilitate linear-time sequence modeling. Notwithstanding their synergistic potential, the integration of these architectures presents a significant challenge, primarily attributable to a fundamental incongruity in their respective positional encoding mechanisms: Transformers rely on explicit Rotary Position Embeddings (RoPE), while SSMs leverage implicit positional representations via convolutions. This divergence often precipitates discontinuities and suboptimal performance. To address this impediment, we propose a unified rotary position embedding (\textbf{\ourRoPE}) methodology, thereby establishing a consistent positional encoding framework for both self-attention and state-space components. Using this \ourRoPE, we introduce \textbf{\model}, a hybrid architecture that coherently integrates the Transformer and SSM layers under this unified positional encoding scheme. At a 4K sequence length, \model exhibits training and inference speeds that are \textbf{42.3\% and 29.5\% faster}, respectively, relative to standard Transformer models. It also delivers higher accuracy: under comparable settings, it surpasses a Transformer baseline by over 4\% on language modeling benchmarks.
\model furthermore scales more effectively:
\model-1.3B gains \textbf{7.22\%} in average accuracy over its 320M version (versus about 6\% gains for equivalent Transformers or SSMs). 
Our results show that unified positional encoding resolves positional incompatibility in hybrid models, enabling efficient, high-performance long-context modeling.
\end{abstract}

\section{Introduction}
\label{sec:introduction}

Effective positional encoding is crucial for sequence modeling in language tasks, as it underpins a model's ability to understand order, perform reasoning, and handle long contexts. Currently, three representative paradigms dominate the landscape of sequence modeling: Transformer-based models (\eg Llama3~\citep{ghahramani2000variational}, state-space models (SSMs, \eg Mamba2~\citep{dao2024ssd}), and hybrid architectures combining both approaches (\eg Jamba~\citep{lieber2024jamba}).
However, as summarized in Table~\ref{tab:TransXSSM-comparison-reduced}, each approach exhibits critical limitations. Transformers utilize explicit positional encodings such as Rotary Position Embedding (RoPE) to effectively handle positional information but suffer from severe computational inefficiencies, especially on long sequences, due to quadratic self-attention complexity~\citep{vaswani2017attention}. 
By contrast, SSM-based models achieve linear computational complexity and high throughput for extremely long sequences but implicitly encode positions, thus significantly limiting their reasoning and few-shot learning abilities. 
Hybrid models (\eg Jamba~\citep{lieber2024jamba}, which combines a Transformer and an Mamba SSM) aim to get the best of both worlds, \ie the reasoning capability of Transformers with the efficiency of SSMs, and have shown promise on extended-context tasks~\citep{waleffe2024empirical}.
However, integrating Transformers and SSMs presents non-trivial challenges, particularly due to fundamental differences in how they encode positional information.
Given these limitations, \textbf{our goal} is to design a hybrid model that effectively combines the powerful positional reasoning capabilities of Transformers with the computational efficiency and scalability of SSMs. 

\begin{table}[t!]
\centering
\caption{Comparison of TransXSSM and related model paradigms across key features.}
\label{tab:TransXSSM-comparison-reduced}
\resizebox{\textwidth}{!}{
\begin{tabular}{cccccccc}
\toprule
\textbf{Models} & \textbf{Mechanism} &
\textbf{Position Encoding}
 & 
 \begin{tabular}[c]{@{}c@{}}\textbf{Unified}\\ \textbf{Encoding}\end{tabular}
 & \begin{tabular}[c]{@{}c@{}}\textbf{Spectrum}\\ \textbf{Continuity}\end{tabular}
 & \begin{tabular}[c]{@{}c@{}}\textbf{Long-Seq}\\ \textbf{Efficiency}\end{tabular}
  & 
  \begin{tabular}[c]{@{}c@{}}\textbf{Training}\\ \textbf{Throughput}\end{tabular}
   & \begin{tabular}[c]{@{}c@{}}\textbf{Inference}\\ \textbf{Speed}\end{tabular} \\ 
\midrule
\begin{tabular}[c]{@{}c@{}}\textbf{Transformer}\\ \textbf{(e.g., Llama3)}\end{tabular}

 & Self-Attention & \begin{tabular}[c]{@{}c@{}}{Rotary}\\ {Position Embedding}\end{tabular} & \textcolor{green}{\cmark} & \textcolor{green}{High} & \textcolor{red}{Low} & \textcolor{red}{Low} & \textcolor{red}{Low} \\ \hline
 \begin{tabular}[c]{@{}c@{}}\textbf{State-Space Models}\\ \textbf{(e.g., Mamba2)}\end{tabular}
  & State-Space & 
  \begin{tabular}[c]{@{}c@{}}{Implicit}\\ {(Conv+Recursion)}\end{tabular}
   & \textcolor{green}{\cmark} & \textcolor{red}{Low} & \textcolor{green}{High} & \textcolor{green}{\textbf{Highest}} & \textcolor{green}{High} \\\hline
   \begin{tabular}[c]{@{}c@{}}\textbf{Hybrid}\\ \textbf{(e.g., Jamba)}\end{tabular}
   & \begin{tabular}[c]{@{}c@{}}{Transformer + }\\ {Mamba}\end{tabular}
    & \textcolor{red}{Separate} & \textcolor{red}{\xmark} & Moderate & Moderate & Moderate & Moderate \\\bottomrule
\textbf{TransXSSM (ours)} & \begin{tabular}[c]{@{}c@{}}{Self-Attention + }\\ {State-Space}\end{tabular}

 & \begin{tabular}[c]{@{}c@{}}{\textbf{\textcolor{green}{Unified}} Rotary}\\ {Position Embedding}\end{tabular}  & \textcolor{green}{\cmark} & \textcolor{green}{High} & \begin{tabular}[c]{@{}c@{}}{\textcolor{green}{High}}\\ {(near-linear)}\end{tabular}   & 
 \begin{tabular}[c]{@{}c@{}}{\textcolor{green}{High}}\\ {(close to SSM)}\end{tabular}
   & \textcolor{green}{High} \\
\bottomrule
\end{tabular}}
\vspace{-1em}
\end{table}

\textbf{Challenges: Positional Encoding Incompatibility and Information Discontinuity.}
Transformer architectures require explicit positional embeddings, such as Rotary Position Embedding (RoPE)~\citep{su2021roformer}, due to their permutation-invariant self-attention mechanism. Conversely, SSMs encode positional information implicitly through convolutional recurrence and internal state dynamics, without explicit positional embeddings. When stacking Transformer and SSM layers naively, this fundamental mismatch in positional encoding methods creates a \textit{positional encoding inconsistency}, causing an \textit{information gap} between module interfaces. Recent work suggests hybrid models can operate without explicit positional encodings by relying solely on SSMs' implicit ordering~\citep{dao2024ssd}. However, this approach can deprive Transformer layers of crucial positional cues, severely impairing performance on tasks that demand precise positional reasoning. Fundamentally, the absence of a unified positional encoding scheme hinders coherent propagation of positional information throughout hybrid models, representing a critical barrier to effectively integrating Transformer and SSM architectures~\citep{patro2024mamba}.

\textbf{Proposed Approach: Unified RoPE and the TransXSSM Architecture.} 
To address the above challenges, we propose a \textit{\underline{unified} \underline{ro}tational \underline{p}osition \underline{e}mbedding} (Unified RoPE) and building a new hybrid architecture around it. Our method extends RoPE, originally created for Transformers to encode absolute positions with complex rotations~\citep{su2021roformer}, to state-space models.
By applying the same rotary positional transformations to the state update signals of an SSM as we do to the query/key vectors of self-attention, we establish a single consistent positional encoding that is shared across both module types. Equipping SSM layers with RoPE-based position embeddings (or adjusting their parameterization to incorporate RoPE) allows positional phase information to flow continuously between attention and SSM layers. The result is a spectrally continuous position representation throughout the hybrid network: no more resets or misalignments of positional phase when transitioning between layers.
%
%
Leveraging Unified RoPE, we further propose the \textbf{\model} architecture, a \textit{Hybrid State-space Attention Model} that integrates self-attention and SSM layers under this unified positional scheme.
As summarized in Table~\ref{tab:TransXSSM-comparison-reduced}, TransXSSM preserves the near-linear runtime, high throughput, and rapid inference typical of state-space models, and, by incorporating explicit relational position encoding together with a lightweight self-attention layer, it matches the deep contextual reasoning capabilities of Transformers.
We make the following contributions:

\begin{itemize}[%
    leftmargin=2em,   %
    labelsep=0.5em,   %
    itemsep=0.25ex     %
]
    \item 
    We propose \textbf{Unified RoPE}, a novel positional encoding mechanism that bridges the gap between self-attention and state-space models (Section~\ref{sec:methods}).

    \item 
    We design \textbf{TransXSSM}, a new hybrid Transformer+SSM model that leverages Unified RoPE for coherence between modules. TransXSSM is carefully structured to combine the strengths of both approaches – it achieves near-linear scaling in sequence length and high throughput (comparable to pure SSMs) while maintaining the robust contextual reasoning ability of Transformers. Key architectural choices (\eg stacking ratio of SSM to attention layers, specialized normalization) enable TransXSSM to maximize efficiency without sacrificing performance (Section~\ref{sec:methods:architecture}).
    

\item \textbf{State-of-the-Art Long-Context Performance}.
We extensively evaluate TransXSSM on language modeling and long-context benchmarks, comparing it against strong baselines (Llama3, Mamba2, Jamba). TransXSSM consistently outperforms these models in both accuracy and speed; notably, our TransXSSM-1.3B model surpasses all baselines by over 2 points on seven diverse tasks. In addition, in a challenging long-context ``needle-in-a-haystack'' retrieval task, TransXSSM demonstrates superior accuracy, underscoring the effectiveness of our unified positional encoding for multi-hop reasoning and deep context modeling (Section~\ref{sec:experiments}).
\end{itemize}

\section{Unified Rotary Position Embedding}
\label{sec:methods}

\subsection{Background: Self-Attention and RoPE}
\label{sec:methods:self_attention}

\textbf{Self-Attention Mechanism.}
Self-attention computes pairwise relevance scores between all tokens in a sequence, enabling long-range dependency modeling~\citep{vaswani2017attention}. For a sequence with query, key, and value representations $Q, K, V$, the basic scaled dot-product attention is: $Y = \operatorname*{softmax}({QK^\top}) \cdot V$.



In causal language modeling, a binary lower-triangular mask $L$ is applied to ensure each position only attends to previous positions (preventing information leakage). This masking restricts the $QK^\top$ matrix to zero out future positions. Many efficient attention variants re-factor or approximate the $QK^\top$ term. For example, linear attention uses kernel feature maps to rewrite $(QK^\top)V$ as $Q\cdot(K^\top V)$, enabling linear complexity. When incorporating the causal mask $L$, the quadratic attention can be implemented recursively. Notably, the masked attention operation can be viewed as a semiseparable matrix application, which connects self-attention to state-space models (SSMs). In fact, \citep{dao2024ssd} show a state-space update can produce the same result as masked attention by an appropriate choice of matrices $C, B$. Formally, one can define a semiseparable matrix $M = L \circ (CB^\top)$ (where “$\circ$” denotes elementwise product) analogous to $M = L \circ (QK^\top)$. Applying this $M$ to an input sequence $X$ (for SSM) or values $V$ (for attention) yields an equivalent result:
$(L \circ QK^\top) \cdot V = (L \circ CB^\top) \cdot X $.



\textbf{Rotary Position Embedding.}
Transformers require explicit positional encoding because the attention operation itself is permutation-invariant. Rotary Position Embedding (RoPE) is a technique that encodes absolute positions as complex rotations applied to query and key vectors. It introduces functions $f_Q$ and $f_K$ such that the inner product of a position-encoded query $q$ at position $m$ and key $k$ at position $n$ depends only on their relative position $(m-n)$. In other words, RoPE achieves an implicit relative position encoding from absolute positions. Formally, for some function $g$:


\begin{align}
    <f_{Q}(Q, m), f_{K}(K, n)> &= g(Q, K, m - n) \label{eq:rope_property}
\end{align}

This means the attention score between token $m$ and token $n$ can be made a function of $m-n$ (their distance), even though each vector is encoded with its absolute position. RoPE was originally proposed for Transformer attention. However, hybrid architectures that intermix Transformers with SSM layers lack a unified positional scheme – Transformers use explicit encodings like RoPE, whereas SSMs typically rely on implicit positional information (\eg via convolutional recurrence). This mismatch can lead to inconsistencies when passing information between layers. We seek to extend RoPE to SSMs so that both module types share the same positional encoding mechanism.


\subsection{Unified Rotary Position Embedding}
\label{sec:methods:unified_rope}

\textbf{Rotations for Hybrid Models.}
We propose a unified rotary position encoding that applies the same rotational embedding to both self-attention (Transformer) and state-space (SSM) components. The key idea is to treat the state-space's recurrence weights analogously to attention's $Q, K$ vectors.
In practice, we define four position-encoding functions $f_Q, f_K, f_C, f_B$ for queries $Q$, keys $K$, and the analogous state update vectors (which we denote $C$ and $B$ for the SSM’s internal update).



Each function multiplies its vector by a complex phase $e^{i,\theta, m}$ or $e^{i,\theta, n}$ determined by the absolute position index. Let $m$ be the position of a ``query-like'' vector ($Q$ or $C$) and $n$ for a ``key-like'' vector ($K$ or $B$). The unified Rotary Position Embedding (Unified RoPE) is defined as:

\begin{equation}
\begin{aligned}
    f_Q(q, m) = q e^{i, m, \theta}, \quad
    f_K(k, n) = k e^{i, n, \theta}, \quad
    f_C(c, m) = c e^{i, m, \theta}, \quad
    f_B(b, n) = b e^{i, n, \theta}
    \label{eq:rope_for_attn_ssd:abs_pos}
\end{aligned}
\end{equation}

Here $\theta$ is a base angular frequency (we will define a set of multiple frequencies $\Theta={\theta_i}$ shortly). Intuitively, multiplying a vector by $e^{i,m\theta}$ rotates it in the complex plane by an angle proportional to its absolute position $m$. Applying the same type of rotation to $Q, K$ in the Transformer, and to $C, B$ in the SSM, produces a consistent positional phase across both modules. This ensures that wherever a sequence goes, \ie attention or SSM, the notion of position is carried by these rotations.

\begin{figure}[!t]
    \centering
    \includegraphics[width=\linewidth]{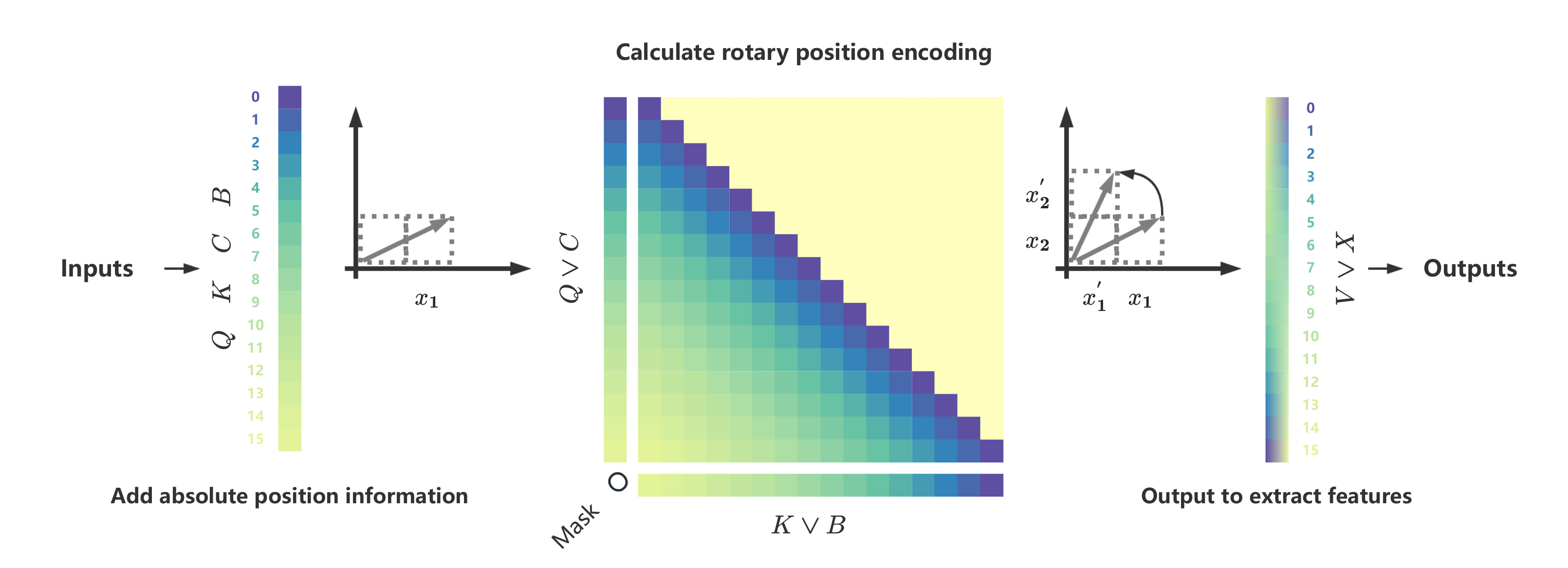}
    \caption{
        \textbf{Rotary Position Embedding Application}.
        Application of Unified RoPE. Input vectors ($Q, C, K, B$) are enriched with absolute positions ($m, n$) via rotation matrices $\mathbb{R}_{\Theta, m}^{d}$ or $\mathbb{R}_{\Theta, n}^{d}$ (Eq.~\eqref{eq:rope_for_attn_ssd:real_form} and~Eq.\eqref{eq:rope_for_attn_ssd:high_dim}), yielding position-encoded vectors. Their inner product captures relative position $m-n$ (Eq.~\eqref{eq:rope_for_attn_ssd:score}). An optional masking step can follow.
    }
    \label{fig:rope_for_attn_ssd}
\end{figure}

\textbf{Real-valued Implementation.} 
To use RoPE in practice, the complex representation (Eq.~\eqref{eq:rope_for_attn_ssd:abs_pos}) is converted to real-valued matrix form. For a 2D vector $[x^{(1)}, x^{(2)}]^\top$, this involves multiplication by a 2D rotation matrix for $f_{\{Q,C\}}$ and $f_{\{K,B\}}$, as shown in Eq.~\eqref{eq:rope_for_attn_ssd:real_form}:


\begin{equation}
\begin{aligned}
    f_{\{Q,C\}}(x_{m}, m) =
    \begin{pmatrix}
        \cos(m \theta) & -\sin(m \theta) \\
        \sin(m \theta) & \cos(m \theta)
    \end{pmatrix}
    \begin{pmatrix}
        x_{m}^{(1)} \\
        x_{m}^{(2)}
    \end{pmatrix}, ~
    f_{\{K,B\}}(x_{n}, n) &=
    \begin{pmatrix}
        \cos(n \theta) & -\sin(n \theta) \\
        \sin(n \theta) & \cos(n \theta)
    \end{pmatrix}
    \begin{pmatrix}
        x_{n}^{(1)} \\
        x_{n}^{(2)}
    \end{pmatrix}
    \label{eq:rope_for_attn_ssd:real_form}
\end{aligned}
\end{equation}

The inner product of these RoPE-modified vectors $f_Q(q_m, m), f_K(k_n, n)$ (or $f_C(c_m, m), f_B(b_n, n)$) yields a score dependent on the relative position $m-n$ via rotation matrix $R_{\Theta, m-n}^{d}$, as shown in Eq.~\eqref{eq:rope_for_attn_ssd:score}. This reflects relative positions for both self-attention and SSMs.
Appendix~\ref{sec:rope_for_ssd} justifies the application of RoPE for the $CB$ term in State-Space Duality.

\begin{equation}
\begin{aligned}
    &attn_{score} = <f_Q(q_m, m), f_K(k_n, n)> = q_m R_{\Theta, m-n}^{d} k_n^{\top} \\
    &ssd_{score} = <f_C(c_m, m), f_B(b_n, n)> = c_m R_{\Theta, m-n}^{d} b_n^{\top}
    \label{eq:rope_for_attn_ssd:score}
\end{aligned}
\end{equation}

In other words, the positional difference $m-n$ is now reflected in the phase of the dot-product. This unifies how positional information is encoded: both the self-attention scores and the SSM update scores can be interpreted through the same relative-position rotation $R^d_{\Theta,m-n}$. As a result, a Transformer layer and an SSM layer will ``see'' positional shifts in the same way, facilitating seamless information flow between the two.

\textbf{Rotation Matrices and Frequencies.}
For dimensions $d > 2$, the vector is partitioned into $d/2$ 2D blocks. Each block is rotated independently (similar to Eq.~\eqref{eq:rope_for_attn_ssd:real_form}) using distinct angles $\theta_i$ from $\Theta$, forming a block-diagonal rotation matrix $R_{\Theta, m}^{d}$ (Eq.~\eqref{eq:rope_for_attn_ssd:high_dim}):

\begin{equation}
R^d_{\Theta,m}
=\bigoplus_{i=0}^{\tfrac d2-1}
\begin{pmatrix}
\cos\bigl(m\theta_i\bigr) & -\sin\bigl(m\theta_i\bigr)\\
\sin\bigl(m\theta_i\bigr) & \cos\bigl(m\theta_i\bigr)
\end{pmatrix},
\quad
\theta_i = 10000^{-2i/d}.
\label{eq:rope_for_attn_ssd:high_dim}
\end{equation}


The set of distinct rotation angles $\Theta = \{\theta_i\}$ for $R_{\Theta, m}^{d}$ (Eq.~\eqref{eq:rope_for_attn_ssd:high_dim}) is defined for $i \in [0, \dots, {\tfrac d2-1}]$ as follows:
$\Theta = \{\theta_i = 10000^{-2i/d}, i \in [0, 1, \dots, {\tfrac d2-1}]\}$.


\textbf{Compatibility with Linear-Time Inference.}
Finally, a causal mask $L$ is applied to the position-encoded score matrices from Eq.~\eqref{eq:rope_for_attn_ssd:score}. These masked scores weight and combine value vectors $V$ (attention) or input vectors $X$ (SSMs) to produce output features $y$, as detailed in Eq.~\eqref{eq:rope_for_attn_ssd:apply_mask}.
Specifically, we multiply elementwise by the mask $L$ (to zero out forbidden positions) and then use the resulting masked scores to weight the corresponding value vectors. For a Transformer attention layer, the values are $V$; for an SSM, the ``values'' are the input sequence $X$ itself. In both cases, we obtain output features $y$ as follows:

\begin{equation}
\begin{aligned}
    y &= (Attn_{score} \circ L) \cdot V, \quad
    y &= (SSD_{score} \circ L) \cdot X
    \label{eq:rope_for_attn_ssd:apply_mask}
\end{aligned}
\end{equation}


Crucially, applying RoPE does not sacrifice the linear-time inference capability of either model. Because RoPE encodes position by a simple rotation of each input vector (Eqs.~\eqref{eq:rope_for_attn_ssd:abs_pos}, \eqref{eq:rope_for_attn_ssd:real_form})  before computing attention or SSM scores, it does not change the asymptotic complexity of those operations. A Transformer with RoPE can still perform auto-regressive generation in linear time by caching past $K,V$ and rotating a new query $q_{m+1}$ on the fly. Similarly, an SSM layer can still update its state recurrently one step at a time, incorporating the rotation $e^{i,\theta, m}$ into its state-update at step $m$. In summary, Unified RoPE preserves the fast inference properties of state-space models and efficient attention mechanisms, while providing a consistent positional encoding throughout a hybrid Transformer+SSM architecture. 

\begin{equation}
    Attn(Q, K, V)_i = \frac{\sum_{j=1}^{i} [\phi(R_{\Theta,i}q_i)] [\varphi(R_{\Theta,j}k_j)]^\top v_j}{\sum_{j=1}^{i} \phi(R_{\Theta,i}q_i) \varphi(R_{\Theta,j}k_j)^\top}, \quad
    SSD(C, B, X)_i = \sum_{j=1}^{i} c_i R_{\Theta, i-j}^{d} b_j^{\top} x_j
    \label{eq:rope_for_attn_ssd:linear_inference}
\end{equation}

Figure~\ref{fig:rope_for_attn_ssd} illustrates the overall process, showing how absolute position rotations lead to relative-position-aware attention maps, which are then masked and used to produce outputs.


\section{TransXSSM Architecture Design}
\label{sec:methods:architecture}

\begin{figure}[!t]
  \centering
  \includegraphics[width=\linewidth]{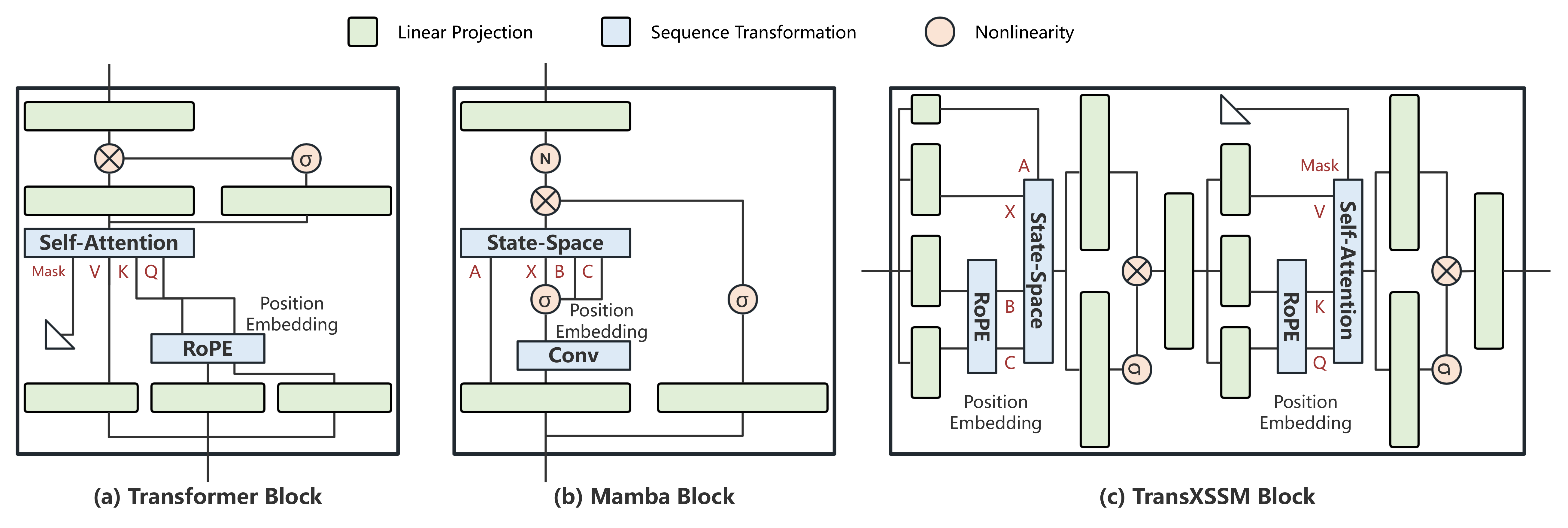}
  \vspace{-2em}
  \caption{
    \textbf{TransXSSM Block}.
    Structure of the TransXSSM block, integrating State-Space (SS), Multi-Layer Perceptron (MLP), and Transformer-style Self-Attention (SA) blocks. All blocks utilize the proposed Unified RoPE.
   }
  \label{fig:TransXSSM_block}
  \vspace{-1em}
\end{figure}

With a unified positional embedding in place, we design an architecture that fully capitalizes on it. TransXSSM (Hybrid Transformer–SSM model) combines the high-level transformer block structure (self-attention + feed-forward network with residual connections) with state-space sequence modeling blocks in a single model. All modules share the same Unified RoPE encoding (as introduced in Section~\ref{sec:methods}), so that positional information is seamlessly understood by every layer (see Figure~\ref{fig:TransXSSM_block}).

\begin{figure}[!t]
  \centering
  \includegraphics[width=\linewidth]{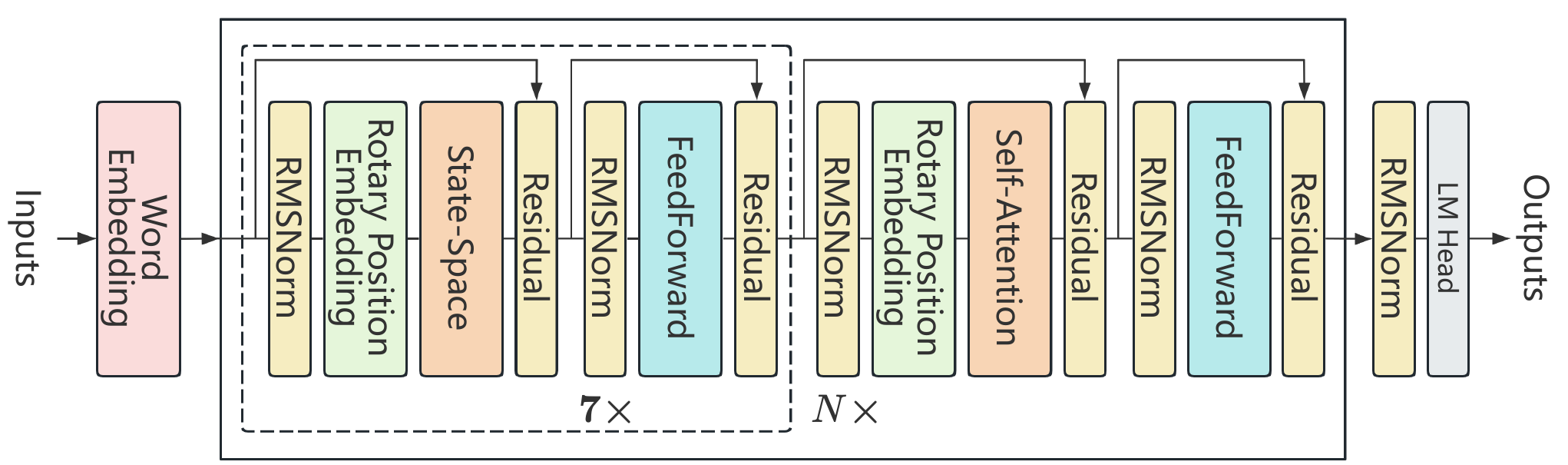}
  \caption{
    \textbf{TransXSSM Language Modeling Architecture}.
    Each input token is first converted to an embedding, then passes through $N$ stacked hybrid modules. Each module contains 7 state-space (SS) layers and 1 self-attention (SA) layer, all employing the unified RoPE positional embedding. Every SS or SA layer is followed by an FFN (feed-forward network), and each pair (SS/SA + FFN) is wrapped with residual connections and RMSNorm normalization. 
  }
  \label{fig:lm_architecture}
  \vspace{-1em}
\end{figure}

\textbf{Overall Architecture.}
For language modeling, TransXSSM follows the standard Transformer architecture outline with modifications to include SSM layers. As shown in Figure~\ref{fig:lm_architecture}, input tokens are first embedded into vectors and then processed by a backbone of $N$ stacked modules. Each module (analogous to a ``layer'') contains multiple State-Space (SS) components and a Self-Attention (SA) component. We adopt a 7:1 ratio of SS to SA blocks per module – that is, each module consists of 7 SSM-based sub-layers followed by 1 Transformer attention sub-layer (this ratio is motivated by prior studies on hybrid models~\citep{blakeman2025nemotron} and our own experiments). Each sub-layer (SS or SA) is followed by a position-wise feed-forward network (FFN) to refine features, and wrapped with residual connections and normalization. We use RMSNorm normalization and add a residual skip connection around each SS/SA block and around its subsequent FFN, which is important for stabilizing training with long sequences (mitigating issues like state collapse observed in long-sequence RNNs~\citep{chen2024stuffed}). Finally, the network output goes through a last normalization and a linear language model head to predict the next-token probabilities.

We highlight four \textbf{key design principles} of TransXSSM.

\textbf{Principle 1: Unified Positional Rotary Encoding.} 
Every SS and SA sub-layer in TransXSSM uses \textit{the same} Unified RoPE as described in Section~\ref{sec:methods:unified_rope}.
Before computing attention scores or state-space updates, the inputs to each layer are endowed with the RoPE phase appropriate for their position. This guarantees that all parts of the model share a common positional frame of reference.

\textbf{Principle 2: Hybrid Module Stacking.}
We intermix a larger number of state-space layers with occasional self-attention layers. In our design, each attention layer is placed after a block of seven SSM layers (7:1 ratio). This strategy, inspired by~\citep{waleffe2024empirical} and validated by our ablation studies, provides an excellent trade-off: the SSM layers efficiently handle the bulk of sequence length (contributing near-linear time complexity and high throughput), while the periodic attention layers inject global context mixing and strong relational reasoning. The result is a model that is both fast on long sequences and capable of complex in-context learning.

\textbf{Principle 3: Feature Refinement via FFNs.}
After each attention or SSM sub-layer, we include a standard Transformer-style feed-forward network. These two-layer FFNs further transform and mix the features. They help ``refine'' the outputs of the SS or SA layers, and are important for maintaining model capacity and expressiveness at each layer of the hybrid model.

\textbf{Principle 4: Stability and Regularization.}
Long sequence training can lead to instabilities such as state-vector drift or ``state collapse'' in SSMs. To ensure robust training, we employ RMSNorm normalization and residual skip connections around every sub-layer (both SS and SA) and its FFN. These measures keep activations bounded and gradients well-behaved over long sequences. Empirically, we found this design choice crucial for training TransXSSM on contexts up to 16K tokens without divergence. The hybrid architecture thus inherits the normalization practices that have proven effective in deep Transformers, applied uniformly across all its components.

By following these principles, TransXSSM effectively unifies two paradigms of sequence modeling. The SSM layers carry the heavy load of long-sequence processing with minimal cost, and the occasional attention layers inject powerful global modeling capabilities – all the while, Unified RoPE ensures that the Transformer and SSM components speak the same positional \textit{language}. In the next section, we empirically demonstrate the benefits of this design, including improved speed, accuracy, and scalability compared to existing approaches.

\section{Experiments}
\label{sec:experiments}

We conduct comprehensive experiments to validate the effectiveness, efficiency, and scalability of the proposed \textit{Unified RoPE} and \textit{TransXSSM} architecture.
We aim to answer the following research questions:
(\textbf{RQ1}) Does Unified RoPE effectively unify position encoding across Transformer and SSM modules, and how does it compare with other positional encodings? (Section~\ref{sec:rq1_unified_rope})
(\textbf{RQ2}) How does TransXSSM compare with state-of-the-art baselines on standard benchmarks? (Section~\ref{sec:rq2_sota_comparison})
(\textbf{RQ3}) Does TransXSSM maintain its advantages at larger model scales? (Section~\ref{sec:rq3_scalability})

\subsection{Effectiveness of Unified RoPE and Comparison with Alternatives}
\label{sec:rq1_unified_rope}

This subsection addresses \textbf{RQ1}. To assess different position encoding methods, we conducted a comparative study. Models with $d_{model}=256$ and sequence length 8192 were trained for 8000 steps. We evaluated Self-Attention, State-Space, and a hybrid State-Space + Self-Attention setup using three schemes: Conv1d + D (1D convolution + dense layer, State-Space only), $a_t$ (recursive, State-Space only), and our proposed \textit{Unified RoPE}. Conv1d + D and $a_t$ are not directly applicable to standard Self-Attention modules. 

The results in Table~\ref{tab:comparative_pe} show that \textbf{Unified RoPE} achieves the best perplexity for both standalone State-Space modules and hybrid configurations when processing long sequences, while maintaining competitive training efficiency. Specifically, in the hybrid State-Space + Self-Attention setup, TransXSSM with RoPE achieved the lowest perplexity (PPL=8.18), with throughput near top-performing state-space models.

Ablation studies further compared RoPE against alternatives (Conv1d + D and $a_t$). The results indicated that these alternatives were ineffective in hybrid scenarios due to incompatibility with self-attention modules, underscoring the necessity of unified positional encoding across modules. This study confirms that Unified RoPE uniquely provides spectral continuity and seamless integration across heterogeneous module types.

\begin{table}[!t]
    \centering
    \caption{
    \textbf{Comparative Position Encoding Performance}.
    Models with $d_{model}=256$ and sequence length 8192 were trained for 8000 steps. RoPE demonstrates superior perplexity and competitive throughput compared to Conv1d + D and $a_t$ for State-Space modules and the hybrid setup. Self-attention modules inherently use RoPE in this comparison.
    }
    \label{tab:comparative_pe}
    \begin{tabular}{@{}lcccccccccc@{}}
    \toprule
    {Modules} & Training Steps & {Conv1d + D} & \sc{$a_t$} & {RoPE} \\
    & & \sc{ppl $\downarrow$} / \sc{s/s $\uparrow$} & \sc{ppl $\downarrow$} / \sc{s/s $\uparrow$} & \sc{ppl $\downarrow$} / \sc{s/s $\uparrow$} \\
    \midrule
    Self-Attention & 8,000 & --- & --- & 8.38 / 6.8 \\
    State-Space & 8,000 & 8.56 / 7.6 & 8.62 / 9.4 & 8.33 / 9.2 \\
    State-Space + Self-Attention & 8,000 & 8.48 / 7.2 & 8.56 / 8.5 & 8.18 / 8.4 \\
    \bottomrule
\end{tabular}
\end{table}

\subsection{Performance Against State-of-the-Art Baselines}
\label{sec:rq2_sota_comparison}

This subsection addresses \textbf{RQ2}. To ensure a fair comparison and demonstrate TransXSSM's advantages in efficiency and effectiveness, we retrained LlaMa3, Mamba2, Jamba, and our proposed TransXSSM. All models were trained under identical conditions, utilizing the Smollm-Corpus dataset, NeoX tokenizer, and consistent key hyperparameters (details in Appendix~\ref{sec:appendix_experimental_setup_for_baseline_comparison}).

\begin{figure}[!t]
    \centering
    \begin{subfigure}{0.48\linewidth}
        \centering
        \includegraphics[width=\linewidth]{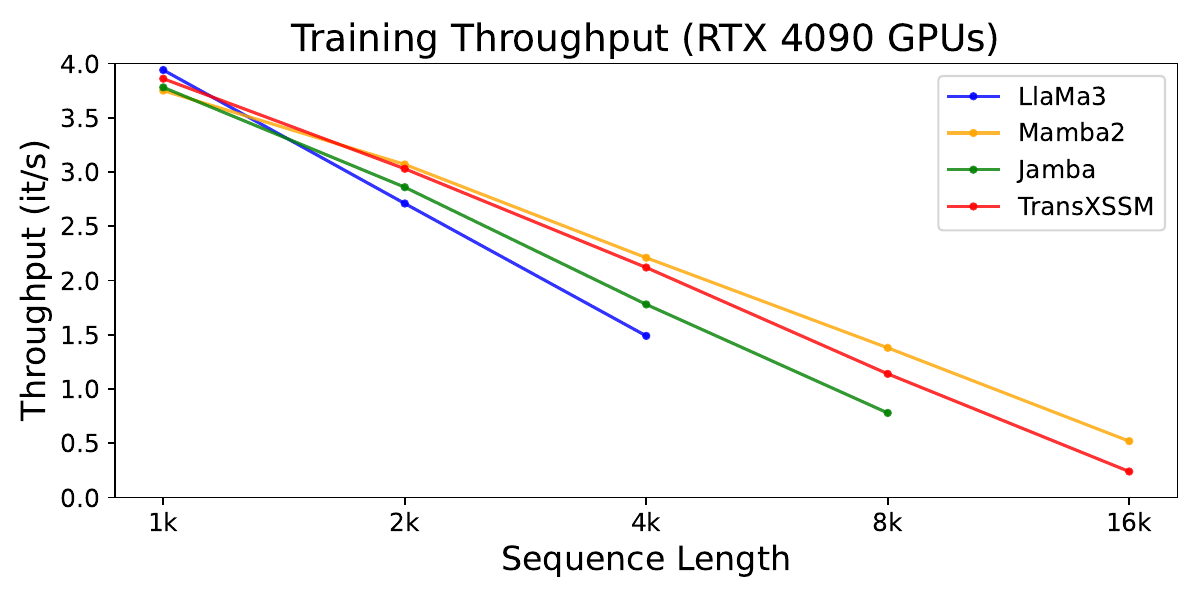}
    \end{subfigure}
    \begin{subfigure}{0.48\linewidth}
        \centering
        \includegraphics[width=\linewidth]{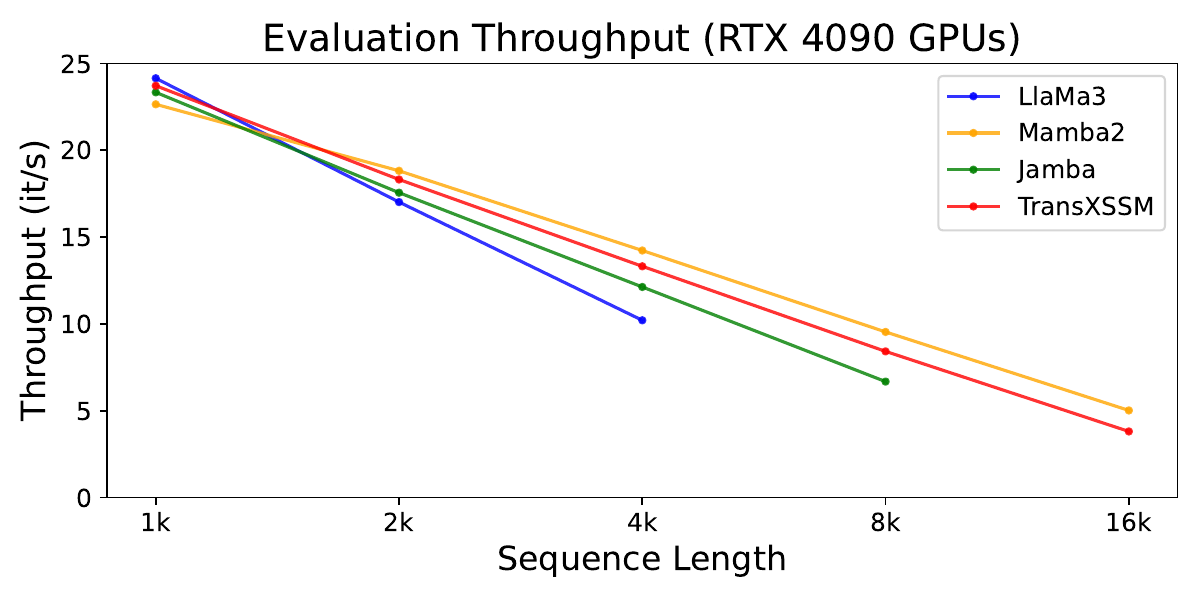}
    \end{subfigure}
    \caption{
        \textbf{Throughput Evaluation}.
        Training (left) and evaluation (right) throughput (iterations/second) for LlaMa3, Mamba2, Jamba, and TransXSSM at the 1.3B parameter scale across varying sequence lengths. TransXSSM surpasses LlaMa3 and Jamba in efficiency, while being slightly less efficient than Mamba2.
    }
    \label{fig:efficient_benchmark}
    \vspace{-1em}
\end{figure}

\textbf{Computational Efficiency and Throughput.} Figure~\ref{fig:efficient_benchmark} shows training and evaluation throughput for 1.3B parameter models across sequence lengths. TransXSSM exhibits superior computational efficiency over LlaMa3 (Self-Attention) and Jamba (hybrid), especially at longer sequence lengths. Although Mamba2 (pure State-Space) is marginally more efficient, TransXSSM's integration of Self-Attention's modeling capabilities justifies this small difference.

\textbf{Performance on Downstream Benchmarks.} Downstream task evaluations (Table~\ref{tab:effective_benchmark}) further substantiate TransXSSM's effectiveness.
\begin{itemize}[%
    leftmargin=2em,   
    labelsep=0.5em,   
    itemsep=0.25ex     
]
    \item \textbf{Cross-Task Stability and Superiority}: TransXSSM shows consistently strong performance across diverse tasks (e.g., MMLU, TriviaQA, PIQA, HellaSwag), often outperforming baselines. This suggests that unified position embedding effectively harmonizes State-Space (efficient long-sequence processing) and Self-Attention (accurate long-dependency capture) strengths.
    \item \textbf{Enhanced Reasoning Capabilities}: TransXSSM leads on tasks requiring commonsense reasoning and contextual understanding (e.g., HellaSwag, PIQA, Winogrande). On Winogrande, the TransXSSM-1.3B outperforms its LlaMa3 counterpart by nearly 6.7 points (62.09 vs 55.40), highlighting the reasoning benefits of our hybrid architecture with unified position encoding.
\end{itemize}

The improvement of TransXSSM over Jamba (another hybrid model) underscores our unified position embedding strategy's importance. While both mix Transformer and SSM blocks, TransXSSM's RoPE-facilitated consistent position representation across components is crucial for coherent contextual understanding and superior performance, vital for tasks needing multi-hop reasoning or precise positional awareness.

\begin{figure}[!t]
  \centering
  \includegraphics[width=\linewidth]{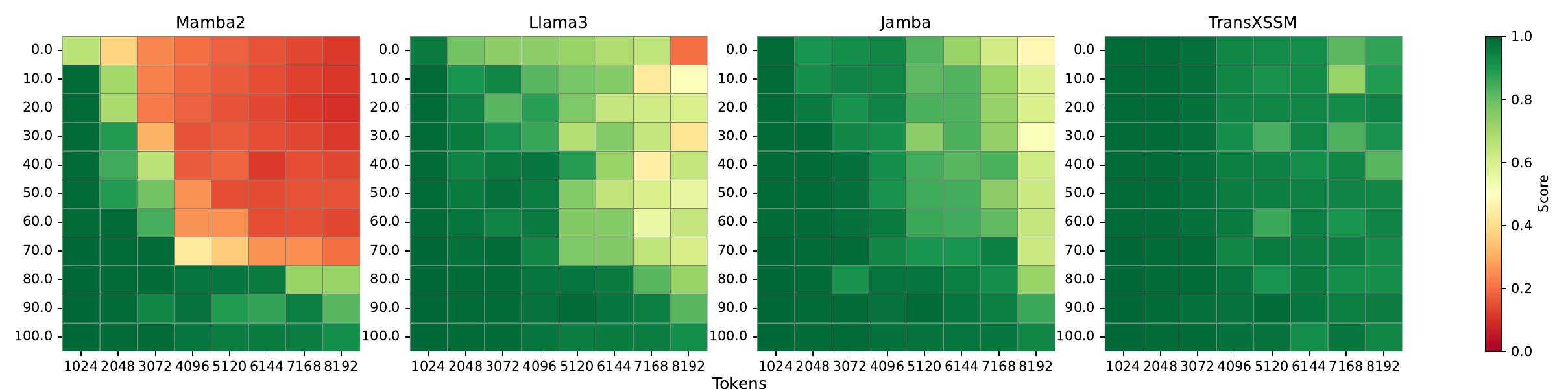}
  \caption{
    \textbf{Needle in a Haystack Performance (1.3B Models)}.
    Performance comparison of LlaMa3, Mamba2, Jamba, and TransXSSM (1.3B scale) on the ``needle in a haystack'' task. TransXSSM, with its unified position encoding, exhibits strong performance.
  }
  \label{fig:needle_in_a_haystack}
\end{figure}

\begin{table}[!t]
    \small
    \setlength{\tabcolsep}{2pt}
    \centering
    \caption{
        \textbf{Downstream Task Performance}.
        Validation results for LlaMa3, Mamba2, Jamba, and TransXSSM on various downstream tasks, trained under identical conditions. Best results per parameter scale are in \textbf{bold}, second best are \underline{underlined}. TransXSSM generally outperforms other models across most tasks and scales.
    }
    \label{tab:effective_benchmark}
    \begin{tabular}{@{}lcccccccccccc@{}}
    \toprule
    \sc{Model} & \sc{MMLU} & \sc{TriviaQA} & \sc{ARC} & \sc{PIQA} & \sc{HellaSwag} & \sc{OBQA} & \sc{Winogrande} & \sc{Avg} \\
    & \sc{acc $\uparrow$} & \sc{qem $\uparrow$} & \sc{acc $\uparrow$} & \sc{acc $\uparrow$} & \sc{acc $\uparrow$} & \sc{acc $\uparrow$} & \sc{acc $\uparrow$} & \\
    \midrule
    LlaMa3-320M & \underline{33.65} & 8.86 & \textbf{51.68} & 71.42 & 52.30 & \underline{37.02} & 53.15 & 43.99 \\
    Mamba2-320M & 33.10 & \underline{9.36} & 50.72 & 70.24 & 48.62 & 35.16 & 54.17 & 43.07 \\
    Jamba-320M & 33.12 & 9.32 & 50.80 & \underline{71.88} & \underline{52.92} & 36.73 & \underline{55.24} & \underline{44.31} \\
    TransXSSM-320M & \textbf{34.45} & \textbf{10.38} & \underline{51.57} & \textbf{73.32} & \textbf{53.79} & \textbf{37.42} & \textbf{55.61} & \textbf{45.22} \\
    \midrule
    LlaMa3-1.3B & \underline{37.86} & 20.66 & \textbf{59.82} & 76.05 & 61.65 & \textbf{41.15} & 55.40 & 50.36 \\
    Mamba2-1.3B & 36.28 & 21.28 & 58.02 & 72.26 & 59.48 & 37.98 & 58.72 & 49.07 \\
    Jamba-1.3B & 37.43 & \underline{21.60} & 59.33 & \underline{76.58} & \underline{62.33} & 40.82 & \underline{59.20} & \underline{51.07} \\
    TransXSSM-1.3B & \textbf{39.08} & \textbf{23.02} & \underline{59.69} & \textbf{78.15} & \textbf{63.63} & \underline{41.12} & \textbf{62.09} & \textbf{52.44} \\
    \bottomrule
    \end{tabular}
    \vspace{-1em}
\end{table}

\textbf{Long-Context Retrieval (Needle-in-a-Haystack Task).} We further evaluated architectures on the ``needle in a haystack'' synthetic retrieval task. This task tests long-context extraction by embedding a ``needle'' (random sentence) in a ``haystack'' (long document) for retrieval. Figure~\ref{fig:needle_in_a_haystack} shows that TransXSSM exhibited strong long-context retrieval, outperforming baselines in this challenging scenario.

\textbf{Task-Specific Strengths and Overall Balance.}  As shown in Table~\ref{tab:effective_benchmark}, it shows pure Transformers (LlaMa3) excel in directed reasoning (e.g., ARC), while pure State-Space models (Mamba2) are strong in direct knowledge extraction (e.g., TriviaQA). TransXSSM, with its 7:1 State-Space to Self-Attention ratio and unified position embedding, retains these respective strengths while achieving a superior overall performance balance, often leading on most tasks.

\subsection{Scalability of TransXSSM Advantages at Larger Model Scales}
\label{sec:rq3_scalability}

To address \textbf{RQ3}, we analyze TransXSSM's advantages when scaling from 320M to 1.3B parameters, comparing its performance progression against baselines (Llama3, Mamba2, Jamba) using configurations from Table~\ref{tab:downstream_evaluation:model} of Appendix~\ref{sec:appendix_experimental_setup_for_baseline_comparison}.

Table~\ref{tab:effective_benchmark} shows that when scaling from 320M to 1.3B parameters, \textbf{TransXSSM's average performance gain (+7.22 points, from 45.22 to 52.44) surpasses that of LlaMa3 (+6.37), Mamba2 (+6.0), and Jamba (+6.76)}. This demonstrates TransXSSM's superior scalability, with its hybrid modeling advantages becoming more pronounced at larger scales. This consistent outperformance during scaling confirms the robustness and effective scalability of TransXSSM's architectural advantages. The unified position embedding provides a solid foundation for effectively fusing different architectural paradigms, leading to a better balance of computational efficiency and model capability at larger scales.

\section{Related Work}
\label{sec:related_work_hybrid}

Hybrid modeling seeks to merge the strengths of different architectures. Recently, architectures combining Transformers and State Space Models (SSMs) have gained attention for their potential to overcome the limitations of unimodal systems.
Early hybrid models explored various integration strategies. For instance, studies combined S4 layers with local attention \cite{zuo2022efficient} or utilized SSM layers preceding Transformer blocks \cite{pilault2023block}, primarily showing promise on small to medium-scale models. However, simple interleaving of Mamba and attention layers \cite{gu2023mamba}, or replacing some attention layers with SSM variants like H3 \cite{fu2022hungry} and Hyena \cite{poli2023hyena}, sometimes struggled to outperform pure Mamba or attention models when scaled, as seen with StripedHyena versus Mistral-7B \cite{gu2023mamba, stripedhyena, jiang2023mistral}. Other attempts included varying the combination order of SSM and self-attention in specific domains like speech recognition \cite{fathullah23_interspeech, saon2023diagonal}, or replacing MLP layers in Transformers with Mamba layers \cite{park2024can}.
The Jamba architecture \citep{lieber2024jamba} marked a significant advancement by achieving a balance between performance, throughput, and memory usage through a carefully designed interleaving ratio of Transformer-Mamba layers, establishing it as a pioneering production-level large-scale attention-SSM hybrid model. Jamba leverages Transformer layers for robust modeling and Mamba layers for efficient long-sequence processing.
Our TransXSSM architecture builds upon these efforts by introducing a novel \textit{unified} Rotary Position Embedding (RoPE) strategy, compatible with both self-attention and state-space duality. This consistent positional understanding across heterogeneous blocks aims to address a critical, often overlooked issue in prior hybrid designs: positional spectrum discontinuity. We posit this as a key factor in our model's enhanced performance and scalability.
\section{Conclusion}
\label{sec:conclusion}

This paper introduces a \textbf{Unified RoPE} method, addressing the core challenge of inconsistent position encoding in hybrid sequence transformation architectures. We theoretically demonstrate the effective application of Rotary Position Embedding to state-space duality algorithms and, based on this, designed the \textbf{TransXSSM} architecture, achieving an efficient fusion of self-attention and state-space models.
Experimental results show that TransXSSM significantly outperforms pure self-attention models, pure state-space models, and other hybrid architectures across various downstream tasks, particularly excelling in knowledge-intensive and reasoning-intensive tasks.
Moreover, TransXSSM's advantages grow with model scale, indicating strong scalability and suggesting it can effectively leverage increased capacity.


\newpage

\appendix

\section{RoPE for SSD}
\label{sec:rope_for_ssd}

\begin{proof}[Proof of Equation~\ref{eq:rope_for_attn_ssd:score}]

by definition, $h_0 = B_0 x_0$.
By induction,

\begin{equation*}
\begin{aligned}
    h_t &= A_t \dots A_1 B_0 x_0 + A_t \dots A_2 B_1 x_1 + \dots + A_t A_{t-1} B_{t-2} x_{t-2} + A_t B_{t-1} x_{t-1} + B_t x_t
    \\&= \sum_{s=0}^t A_{t:s}^\times B_s x_s
\end{aligned}
\end{equation*}

Multiplying by $C_t$ to produce $y_t$, and vectorizing the equation to $t \in [\mathtt{T}]$ ($\mathtt{T}$ is the sequence length), we derive the matrix transformation form of SSD.

\begin{equation*}
\begin{aligned}
    y_t &= \sum_{s=0}^t C_t^{\top} A_{t:s}^\times B_s x_s
    \\
    y &= \mathsf{SSD}(A, B, C)(x) = Mx
    \\
    M_{ji} &:= C_j^{\top} A_{j} \cdots A_{i+1} B_{i}
\end{aligned}
\end{equation*}

Then the matrix form of SSD is represented using SSS (Sequentially Semiseparable) as $M = \mathsf{SSS}(A, B, C)$, where $M_{ji} = C_j^{\top} A_{j:i} B_i$, and then considering $A$ is just a scalar, rearranged as:

\begin{equation*}
    \begin{aligned}
    M_{ji} = A_{j:i} \cdot (C_j^{\top}B_i)
    \end{aligned}
\end{equation*}

Vectorized as:

\begin{equation*}
    \begin{aligned}
    L &:= \mathsf{1SS}(a) \\
    M &= L \circ (C B^{\top}) \\
    \end{aligned}
\end{equation*}

Finally, it is proved that the matrix transformation form of SSD is equivalent to Attention $(L \circ QK^\top) \cdot V = (L \circ CB^\top) \cdot X$.

Now we have enough theoretical support to give rotational positional encoding to the $C$ and $B$ matrices in SSD.

\begin{equation*}
    \begin{aligned}
        C_{m} &= f_{C} (x_{m}, m) \\
        B_{n} &= f_{B} (x_{n}, n)
    \end{aligned}
\end{equation*}

$C_{m}$ represents the output weight matrix of the $m$-th token corresponding to the word vector $x_{m}$ integrated with the position information $m$,
$B_{n}$ represents the input weight matrix of the $n$-th token corresponding to the word vector $x_{n}$ integrated with the position information $n$.

To utilize the relative positional information between tokens,
we assume that the inner product operation between the $C_{m}$ vector and the $B_{n}$ vector can be represented by a function $g$,
where the input of the function $g$ is the word embedding vectors $x_{m}$ and $x_{n}$,
and their relative positional information $m - n$,
the inner product of $C_{m}$ and $B_{n}$ and their relative positional information $m - n$ is defined as

\begin{equation*}
    \begin{aligned}
    <f_{C}(x_{m}, m), f_{B}(x_{n}, n)> = g(x_{m}, x_{n}, m - n)
    \end{aligned}
\end{equation*}

Now, assuming the word embedding vector dimension is $d = 2$,
we have $f_{C}(x_{m}, n) = (W_{C} x_{m})e^{i m \theta}$,
for the first half of the formula $W_{C} x_{m}$,
we know that $W_{C}$ is a two-dimensional matrix,
$x_{m}$ is a two-dimensional vector,
the result of the multiplication is naturally a two-dimensional vector,
represented by $C_{m}$

\begin{equation*}
    \begin{aligned}
    C_{m} &= 
        \begin{bmatrix} 
            C_{m}^{(1)} \\
            C_{m}^{(2)}
        \end{bmatrix}
    = W_{C} x_{m} = 
        \begin{bmatrix} 
            W_{C}^{(11)} & W_{C}^{(12)} \\
            W_{C}^{(21)} & W_{C}^{(22)}
        \end{bmatrix}
        \begin{bmatrix} 
            x_{m}^{(1)} \\
            x_{m}^{(2)}
        \end{bmatrix}
\end{aligned}
\end{equation*}

For the second half $e^{i m \theta}$,
according to Euler's formula $e^{i x} = \cos(x) + i \sin(x)$,
we have

\begin{equation*}
    \begin{aligned}
    e^{i m \theta} &= \cos(m \theta) + i \sin(m \theta)
    \end{aligned}
\end{equation*}

We know

\begin{equation*}
    \begin{aligned}
    f_{C}(x_{m}, m) &= (W_{C} x_{m})e^{i m \theta} = C_{m}e^{i m \theta}
    \end{aligned}
\end{equation*}

$C_{m}$ is represented in complex form,

\begin{equation*}
    \begin{aligned}
    C_{m} &=
        \begin{bmatrix} 
            C_{m}^{(1)}, C_{m}^{(2)}
        \end{bmatrix}
    = 
        \begin{bmatrix} 
            C_{m}^{(1)} + i C_{m}^{(2)}
        \end{bmatrix}
    \end{aligned}
\end{equation*}

Thus,

\begin{equation*}
    \begin{aligned}
    f_{C}(x_{m}, m) &= C_{m}e^{i m \theta} = 
        \begin{bmatrix} 
            C_{m}^{(1)} + i C_{m}^{(2)}
        \end{bmatrix} e^{i m \theta}
    \end{aligned}
\end{equation*}

According to the above derivation,
we know that $f_{C}(x_{m}, m)$ is the product of two complex numbers,

\begin{equation*}
    \begin{aligned}
    f_{C}(x_{m}, m) &= C_{m}e^{i m \theta} = 
        \begin{bmatrix} 
            C_{m}^{(1)} + i C_{m}^{(2)}
        \end{bmatrix} \times (\cos(m \theta) + i \sin(m \theta))
    \end{aligned}
\end{equation*}

Considering the following two formulas about complex numbers

\begin{equation*}
    \begin{aligned}
    (a + i b) \times (c + i d) &= ac + i bc + i ad + i^2 bd = (ac - bd) + i (bc + ad) \\
    i^2 &= -1
    \end{aligned}
\end{equation*}

We have

\begin{equation*}
    \begin{aligned}
    C_{m}e^{i m \theta} &= 
        \begin{bmatrix} 
            C_{m}^{(1)} + i C_{m}^{(2)}
        \end{bmatrix} \times (\cos(m \theta) + i \sin(m \theta)) \\
        &= 
        \begin{bmatrix} 
            C_{m}^{(1)} \cos(m \theta) - C_{m}^{(2)} \sin(m \theta)
        \end{bmatrix} + i
        \begin{bmatrix} 
            C_{m}^{(2)} \cos(m \theta) + C_{m}^{(1)} \sin(m \theta)
        \end{bmatrix}
    \end{aligned}
\end{equation*}

Expressing this result as a real vector,

\begin{equation*}
    \begin{aligned}
    C_{m}e^{i m \theta} &= 
        \begin{bmatrix} 
            C_{m}^{(1)} \cos(m \theta) - C_{m}^{(2)} \sin(m \theta),
            C_{m}^{(2)} \cos(m \theta) + C_{m}^{(1)} \sin(m \theta)
        \end{bmatrix}
    \end{aligned}
\end{equation*}

Therefore, $C_{m}$ multiplied by a rotation matrix is obtained.

\begin{equation*}
    \begin{aligned}
    f_{C}(x_{m}, m) &= (W_{C} x_{m})e^{i m \theta} = C_{m}e^{i m \theta} \\
    &=
    \begin{bmatrix} 
        C_{m}^{(1)} \cos(m \theta) - C_{m}^{(2)} \sin(m \theta),
        C_{m}^{(2)} \cos(m \theta) + C_{m}^{(1)} \sin(m \theta)
    \end{bmatrix} \\
    &=
    \begin{bmatrix} 
        \cos(m \theta) & -\sin(m \theta) \\
        \sin(m \theta) & \cos(m \theta)
    \end{bmatrix}
    \begin{bmatrix} 
        C_{m}^{(1)} \\
        C_{m}^{(2)}
    \end{bmatrix}
    \end{aligned}
\end{equation*}

Similarly, $B_{n}$ vector can be obtained

\begin{equation*}
    \begin{aligned}
    f_{B}(x_{n}, n) &= (W_{B} x_{n})e^{i n \theta} = B_{n}e^{i n \theta} \\
    &=
    \begin{bmatrix} 
        B_{n}^{(1)} \cos(n \theta) - B_{n}^{(2)} \sin(n \theta),
        B_{n}^{(2)} \cos(n \theta) + B_{n}^{(1)} \sin(n \theta)
    \end{bmatrix} \\
    &=
    \begin{bmatrix} 
        \cos(n \theta) & -\sin(n \theta) \\
        \sin(n \theta) & \cos(n \theta)
    \end{bmatrix}
    \begin{bmatrix} 
        B_{n}^{(1)} \\
        B_{n}^{(2)}
    \end{bmatrix}
    \end{aligned}
\end{equation*}

The function $g$ can be represented as

\begin{equation*}
    \begin{aligned}
    g(x_{m}, x_{n}, m - n) &= R 
        \begin{bmatrix} 
            (W_{C} x_{m})(W_{B} x_{n})^*e^{i (m - n) \theta}
        \end{bmatrix}
    \end{aligned}
\end{equation*}

where $R$ represents the real part of the complex number $x$,
$(W_{C} x_{m})(W_{B} x_{n})^*$ represents the conjugate of the product of two complex numbers.
Considering

\begin{equation*}
    \begin{aligned}
    z &= a + i b \\
    z^* &= a - i b
    \end{aligned}
\end{equation*}

we have

\begin{equation*}
    \begin{aligned}
    W_{C} x_{m} &= C_{m} = C_{m}^{(1)} + i C_{m}^{(2)} \\
    W_{B} x_{n} &= B_{n} = B_{n}^{(1)} + i B_{n}^{(2)} \\
    (W_{B} x_{n})^* &= B_{n}^* = B_{n}^{(1)} - i B_{n}^{(2)} \\
    e^{i (m - n) \theta} &= \cos((m - n) \theta) + i \sin((m - n) \theta)
    \end{aligned}
\end{equation*}

We now want to prove that

\begin{equation*}
    \begin{aligned}
    &g(x_{m}, x_{n}, m - n) \\
    &= R 
        \begin{bmatrix} 
            (W_{C} x_{m})(W_{B} x_{n})^*e^{i (m - n) \theta}
        \end{bmatrix} \\
    &= R
        \begin{bmatrix} 
            (C_{m}^{(1)} + i C_{m}^{(2)})(B_{n}^{(1)} - i B_{n}^{(2)})(\cos((m - n) \theta) + i \sin((m - n) \theta))
        \end{bmatrix} \\
    &= R
        \begin{bmatrix} 
            ((C_{m}^{(1)}B_{n}^{(1)} + C_{m}^{(2)}B_{n}^{(2)}) + i (C_{m}^{(2)}B_{n}^{(1)} - C_{m}^{(1)}B_{n}^{(2)}))(\cos((m - n) \theta) + i \sin((m - n) \theta))
        \end{bmatrix} \\
    &= (C_{m}^{(1)}B_{n}^{(1)} + C_{m}^{(2)}B_{n}^{(2)})\cos((m - n) \theta) - (C_{m}^{(2)}B_{n}^{(1)} - C_{m}^{(1)}B_{n}^{(2)})\sin((m - n) \theta)
    \end{aligned}
\end{equation*}

Recalling the vectorized form of SSD,
the $C$ vector at position $m$ and the $B$ vector at position $n$ will perform an inner product operation,
that is,

\begin{equation*}
    \begin{aligned}
    f_{C}(x_{m}, m) &= 
        \begin{bmatrix} 
            C_{m}^{(1)} \cos(m \theta) - C_{m}^{(2)} \sin(m \theta),
            C_{m}^{(2)} \cos(m \theta) + C_{m}^{(1)} \sin(m \theta)
        \end{bmatrix} \\
    f_{B}(x_{n}, n) &=
        \begin{bmatrix} 
            B_{n}^{(1)} \cos(n \theta) - B_{n}^{(2)} \sin(n \theta),
            B_{n}^{(2)} \cos(n \theta) + B_{n}^{(1)} \sin(n \theta)
        \end{bmatrix}
    \end{aligned}
\end{equation*}

We have

\begin{equation*}
    \begin{aligned}
    <f_{C}(x_{m}, m), f_{B}(x_{n}, n)> &= 
        \begin{bmatrix} 
            C_{m}^{(1)} \cos(m \theta) - C_{m}^{(2)} \sin(m \theta)
        \end{bmatrix}
        \begin{bmatrix} 
            B_{n}^{(1)} \cos(n \theta) - B_{n}^{(2)} \sin(n \theta)
        \end{bmatrix} \\
        &+
        \begin{bmatrix} 
            C_{m}^{(2)} \cos(m \theta) + C_{m}^{(1)} \sin(m \theta)
        \end{bmatrix}
        \begin{bmatrix} 
            B_{n}^{(2)} \cos(n \theta) + B_{n}^{(1)} \sin(n \theta)
        \end{bmatrix} \\
        &= C_{m}^{(1)} \cos(m \theta) B_{n}^{(1)} \cos(n \theta) - C_{m}^{(1)} \cos(m \theta) B_{n}^{(2)} \sin(n \theta) \\
        &- C_{m}^{(2)} \sin(m \theta) B_{n}^{(1)} \cos(n \theta) + C_{m}^{(2)} \sin(m \theta) B_{n}^{(2)} \sin(n \theta) \\
        &+ C_{m}^{(2)} \cos(m \theta) B_{n}^{(2)} \cos(n \theta) + C_{m}^{(2)} \cos(m \theta) B_{n}^{(1)} \sin(n \theta) \\
        &+ C_{m}^{(1)} \sin(m \theta) B_{n}^{(2)} \cos(n \theta) + C_{m}^{(1)} \sin(m \theta) B_{n}^{(1)} \sin(n \theta)
    \end{aligned}
\end{equation*}

Considering

\begin{equation*}
    \begin{aligned}
    \sin(a + b) &= \sin(a)\cos(b) + \cos(a)\sin(b) \\
    \sin(a - b) &= \sin(a)\cos(b) - \cos(a)\sin(b) \\
    \cos(a + b) &= \cos(a)\cos(b) - \sin(a)\sin(b) \\
    \cos(a - b) &= \cos(a)\cos(b) + \sin(a)\sin(b)
    \end{aligned}
\end{equation*}

We have

\begin{equation*}
    \begin{aligned}
        &<f_{C}(x_{m}, m), f_{B}(x_{n}, n)> \\
        &= C_{m}^{(1)} B_{n}^{(1)} (\cos(m \theta) \cos(n \theta) + \sin(m \theta) \sin(n \theta)) \\
        &+ C_{m}^{(1)} B_{n}^{(2)} (-\cos(m \theta) \sin(n \theta) + \sin(m \theta) \cos(n \theta)) \\
        &+ C_{m}^{(2)} B_{n}^{(1)} (-\sin(m \theta) \cos(n \theta) + \cos(m \theta) \sin(n \theta)) \\
        &+ C_{m}^{(2)} B_{n}^{(2)} (\sin(m \theta) \sin(n \theta) + \cos(m \theta) \cos(n \theta)) \\
        &= C_{m}^{(1)} B_{n}^{(1)} \cos((m - n) \theta) + C_{m}^{(1)} B_{n}^{(2)} \sin((m - n) \theta) \\
        &- C_{m}^{(2)} B_{n}^{(1)} \sin((m - n) \theta) + C_{m}^{(2)} B_{n}^{(2)} \cos((m - n) \theta) \\
        &= (C_{m}^{(1)} B_{n}^{(1)} + C_{m}^{(2)} B_{n}^{(2)})\cos((m - n) \theta) + (C_{m}^{(1)} B_{n}^{(2)} - C_{m}^{(2)} B_{n}^{(1)})\sin((m - n) \theta) \\
        &= (C_{m}^{(1)}B_{n}^{(1)} + C_{m}^{(2)}B_{n}^{(2)})\cos((m - n) \theta) - (C_{m}^{(2)}B_{n}^{(1)} - C_{m}^{(1)}B_{n}^{(2)})\sin((m - n) \theta) \\
        &= g(x_{m}, x_{n}, m - n)
    \end{aligned}
\end{equation*}

It is proved that the inner product of the $C$ vector at position $m$ and the $B$ vector at position $n$ is the function $g$.

Finally, using the matrix-vector multiplication form

\begin{equation*}
    \begin{aligned}
    &<f_{C}(x_{m}, m), f_{B}(x_{n}, n)> \\
    &= 
        \begin{bmatrix} 
            \begin{bmatrix} 
                \cos(m \theta) & -\sin(m \theta) \\
                \sin(m \theta) & \cos(m \theta)
            \end{bmatrix}
            \begin{bmatrix} 
                C_{m}^{(1)} \\
                C_{m}^{(2)}
            \end{bmatrix}
        \end{bmatrix}^{T}
        \begin{bmatrix} 
            \begin{bmatrix} 
                \cos(n \theta) & -\sin(n \theta) \\
                \sin(n \theta) & \cos(n \theta)
            \end{bmatrix}
            \begin{bmatrix} 
                B_{n}^{(1)} \\
                B_{n}^{(2)}
            \end{bmatrix}
        \end{bmatrix} \\
        &= 
        \begin{bmatrix} 
            C_{m}^{(1)} & C_{m}^{(2)}
        \end{bmatrix}
        \begin{bmatrix} 
            \cos(m \theta) & \sin(m \theta) \\
            -\sin(m \theta) & \cos(m \theta)
        \end{bmatrix}
        \begin{bmatrix} 
            \cos(n \theta) & -\sin(n \theta) \\
            \sin(n \theta) & \cos(n \theta)
        \end{bmatrix}
        \begin{bmatrix} 
            B_{n}^{(1)} \\
            B_{n}^{(2)}
        \end{bmatrix} \\
    \end{aligned}
\end{equation*}

Expanding the product of the two rotary matrices, we have

\begin{equation*}
    \begin{aligned}
    \begin{bmatrix} 
        \cos(m \theta) \cos(n \theta) + \sin(m \theta) \sin(n \theta) & -\cos(m \theta) \sin(n \theta) + \sin(m \theta) \cos(n \theta) \\
        -\sin(m \theta) \cos(n \theta) + \cos(m \theta) \sin(n \theta) & \sin(m \theta) \sin(n \theta) + \cos(m \theta) \cos(n \theta)
    \end{bmatrix}
    \end{aligned}
\end{equation*}

Finally, we get

\begin{equation*}
    \begin{aligned}
    <f_{C}(x_{m}, m), f_{B}(x_{n}, n)> &= 
        \begin{bmatrix} 
            C_{m}^{(1)} & C_{m}^{(2)}
        \end{bmatrix}
        \begin{bmatrix} 
            \cos((m - n) \theta) & -\sin((m - n) \theta) \\
            \sin((m - n) \theta) & \cos((m - n) \theta)
        \end{bmatrix}
        \begin{bmatrix} 
            B_{n}^{(1)} \\
            B_{n}^{(2)}
        \end{bmatrix}
    \end{aligned}
\end{equation*}

The above derivation is only for the case of word embedding dimension $d = 2$,
when $d > 2$, the two-dimensional case can be extended to any dimension as follows

\begin{equation*}
    \begin{aligned}
    f_{\{C, B\}}(x_{m}, m) &= R_{\Theta, m}^{d} W_{\{C, B\}} x_{m}
    \end{aligned}
\end{equation*}

The inner product satisfies linearity,
so for any even-dimensional RoPE, we can represent it as a concatenation of the two-dimensional case,
that is, grouping the elements of the word embedding vector in pairs

\begin{equation*}
    \begin{aligned}
    R_{\Theta, m}^{d} = \begin{bmatrix} 
        \cos m \theta_0 & -sin m \theta_0 & 0 & 0 & \dots & 0 & 0 \\
        \sin m \theta_0 & \cos m \theta_0 & 0 & 0 & \dots & 0 & 0 \\
        0 & 0 & \cos m \theta_1 & -sin m \theta_1 & \dots & 0 & 0 \\
        0 & 0 & \sin m \theta_1 & \cos m \theta_1 & \dots & 0 & 0 \\
        \vdots & \vdots & \vdots & \vdots & \ddots & \vdots & \vdots \\
        0 & 0 & 0 & 0 & \dots & \cos m \theta_{d/2} & -sin m \theta_{d/2-1} \\
        0 & 0 & 0 & 0 & \dots & \sin m \theta_{d/2} & \cos m \theta_{d/2-1} \\
    \end{bmatrix}
    \end{aligned}
\end{equation*}

Each group applies the same rotation operation and the rotation angle of each group is calculated as follows:

\begin{equation*}
    \begin{aligned}
    \Theta &= \{\theta_i = 10000^{-2(i - 1) / d}, i \in [1, 2, \dots, d / 2]\}
    \end{aligned}
\end{equation*}
  
RoPE is a kind of relative positional encoding, and relative positional encoding is a special form of Toeplitz matrix

\begin{equation*}
    \begin{aligned}
    \begin{bmatrix} 
        0 & & & & & & \\
        1 & 0 & & & & \\
        2 & 1 & 0 & & \\
        3 & 2 & 1 & 0 \\
        \vdots & \ddots & \ddots & \ddots & \ddots \\
        n-1 & n-2 & n-3 & \dots & 1 & 0
    \end{bmatrix}
    \end{aligned}
\end{equation*}

We can know that the position distribution after RoPE is unbalanced, 0 appears most frequently as the low bit, and n-1 appears least frequently as the high bit, which also leads to a problem of RoPE, its high bit is not sufficiently trained, and the generalization ability is not as good as the low bit. We take the average value of the effective sequence length of the training data as the base, for the sequence length greater than the base, we have

\begin{equation*}
    \begin{aligned}
    C \times \max(1, \log_{base} n) \\
    \end{aligned}
\end{equation*}

The part of the sequence length within the base is not affected, and the part of the sequence length greater than the base is expanded according to the ratio of $\log_{base} n$, so that the problem of insufficient generalization ability of the high bit can be solved.

\end{proof}

\newpage
\section{Additional Experimental Details}
\label{sec:appendix_additional_experimental_details}

\subsection{Experimental Setup for Baseline Comparison}
\label{sec:appendix_experimental_setup_for_baseline_comparison}

\paragraph{Model Architectures Settings.}
To ensure a fair comparison and avoid discrepancies from varying training data, we retrained four distinct model architectures: Llama3~\citep{grattafiori2024llama3herdmodels} (Self-Attention based), Mamba2~\citep{dao2024ssd} (State-Space based), Jamba~\citep{lieber2024jamba} (a hybrid of Transformer and Mamba2 blocks), and our proposed TransXSSM. All models were trained from scratch on the Smollm-Corpus~\citep{benallal2024smollmcorpus} dataset using the NeoX tokenizer~\citep{black2022gpt}.

\paragraph{Model Scales Settings.}
We experimented with two model scales, 320M and 1.3B parameters, detailed in Table~\ref{tab:downstream_evaluation:model}. For the 320M scale, models have $d_{model}=768$, 24 layers, 12 attention heads, a learning rate of 3e-4, and a batch size of 1M tokens. For the 1.3B scale, $d_{model}=2048$, 24 layers, 32 attention heads, a learning rate of 2e-4, and a batch size of 2M tokens. State-Space components within Mamba2, Jamba, and TransXSSM uniformly use $d_{state}=128$ and $chunk\_len=256$.

\paragraph{Model Training Settings.}
Training was conducted using Nvidia's open-source PyTorch container (version 24.2), compatible with the mamba-ssm library's CUDA kernels, and managed by the Trainer class from the Transformers library~\citep{wolf-etal-2020-transformers}. We employed the AdamW optimizer ($\beta_1=0.9, \beta_2=0.999, weight\_decay=0.01$), with a linear warmup for $10\%$ of total steps followed by a cosine decay to $10\%$ of the initial learning rate. No bias terms were used, and LayerNorm was replaced by RMSNorm.

\begin{table}[!h] 
    \centering
    \setlength{\tabcolsep}{2pt}
    \caption{
        \textbf{Model Parameters}.
        Key hyperparameters for Llama3, Mamba2, Jamba, and TransXSSM models at 320M and 1.3B scales. Parameters were aligned where possible to ensure comparable total and active parameter counts.
    }
    \label{tab:downstream_evaluation:model}
    \begin{tabular}{@{}lcccccccc@{}} 
    \toprule
    \sc{Model} & $d_{model}$ & $n_{layers}$ & $n_{heads}$ & $d_{state}$ & chunk\_len & \sc{Leaning Rate} & \sc{Batch Size} \\
    \midrule
    LlaMa3-320M & 768 & 24 & 12 & --- & --- & 3e-4 & 1M tokens \\
    Mamba2-320M & 768 & 24 & 12 & 128 & 256 & 3e-4 & 1M tokens \\
    Jamba-320M & 768 & 24 & 12 & 128 & 256 & 3e-4 & 1M tokens \\
    TransXSSM-320M & 768 & 24 & 12 & 128 & 256 & 3e-4 & 1M tokens \\
    \midrule
    LlaMa3-1.3B & 2048 & 24 & 32 & --- & --- & 2e-4 & 2M tokens \\
    Mamba2-1.3B & 2048 & 24 & 32 & 128 & 256 & 2e-4 & 2M tokens \\
    Jamba-1.3B & 2048 & 24 & 32 & 128 & 256 & 2e-4 & 2M tokens \\
    TransXSSM-1.3B & 2048 & 24 & 32 & 128 & 256 & 2e-4 & 2M tokens \\
    \bottomrule
    \end{tabular}
\end{table}

\paragraph{Downstream Tasks Settings.}
Downstream task evaluation utilized the EleutherAI LM Evaluation Harness~\citep{eval-harness}. The benchmark suite included MMLU~\citep{hendrycks2021measuring}, TriviaQA~\citep{m2017triviaqa}, ARC~\citep{clark2018think}, PIQA~\citep{bisk2020piqa}, HellaSwag~\citep{zellers2019hellaswag}, OBQA~\citep{mihaylov2018can}, and Winogrande~\citep{sakaguchi2021winogrande}. Consistent evaluation settings were applied across all 320M and 1.3B models to ensure fair and comparable results.
\newpage

{

\small

\bibliographystyle{IEEEtran} \bibliography{refs/biblio}

}
\end{document}